\documentclass{article}

\usepackage{arxiv}
\usepackage[utf8]{inputenc}
\usepackage[T1]{fontenc}
\usepackage{hyperref}
\usepackage{url}
\usepackage{graphicx}
\usepackage{amsmath,amsfonts}
\usepackage{booktabs}
\usepackage{multirow}
\usepackage{microtype}
\usepackage{enumitem}
\usepackage{xcolor}
\usepackage{cite}
\usepackage{tikz}
\usepackage[ruled,vlined]{algorithm2e}
\usetikzlibrary{arrows.meta, positioning, calc}

\title{ThirdEye: Cue‑Aware Monocular Depth Estimation via Brain‑Inspired Multi‑Stage Fusion}

\author{Calin Teodor Ioan\\
DartLabs\\
\texttt{teodor@dart-labs.com}}

\begin{document}
\maketitle

\begin{abstract}
Monocular depth estimation methods traditionally rely on deep learning models to implicitly understand complex depth features directly from RGB pixels. However, this approach often neglects explicit monocular cues that the human visual system naturally exploits—such as occlusion boundaries, shading patterns, and perspective layouts. Instead of expecting models to learn these cues independently, we propose \textbf{ThirdEye}, a cue-aware pipeline that explicitly "spoon-feeds" carefully extracted monocular cues from specialized, pre-trained, and frozen networks. Inspired by biological vision, ThirdEye fuses these explicitly provided cues through a hierarchical, cortically-inspired structure (V1→V2→V3), integrated with a key–value working-memory system that dynamically weighs cues based on reliability. The depth estimation is finalized by an adaptive-bins transformer head, yielding high-resolution disparity maps. This modular approach allows ThirdEye to benefit significantly from external supervision, requiring minimal fine-tuning. This extended version elaborates further on the architectural rationale, neuroscientific foundations, and a detailed experimental framework, with quantitative analyses forthcoming in subsequent revisions.
\end{abstract}

\section{Introduction}
Monocular depth estimation (MDE) underpins autonomous driving, augmented reality (AR) and robotics. Classical deep networks regress raw pixels to depth but often neglect explicit physical cues, resulting in opaque reasoning \cite{Eigen2014}.  By contrast, the primate visual system integrates multiple monocular cues (shading, occlusion, perspective) and maintains hypotheses via top‑down feedback. Psychophysical and neuro‑imaging studies show that early visual cortex (V1–V3) encodes depth‑relevant information and exhibits persistent activity during memory tasks \cite{natureWM2022,feedback2024}.

\textbf{ThirdEye} emulates this biological strategy. It combines off‑the‑shelf cue extractors—HED edges \cite{xie2015}, SDPS‑Net normals \cite{Chen2019SDPS} and HorizonNet layout \cite{Sun2019Horizon}—with a cortical‑style fusion hierarchy and an explicit key–value working‑memory bank.  By freezing the specialists, the model separates \emph{where} cues originate from \emph{how} they are fused, enabling plug‑and‑play upgrades and continual‑learning extensions.

\paragraph{Contributions.}
\begin{enumerate}[leftmargin=2em]
  \item We formulate a \emph{cue‑aware}, plug‑and‑play depth pipeline integrating frozen, publicly available specialists.
  \item We propose a \emph{brain‑inspired fusion architecture} featuring a lightweight working‑memory bank realised as key–value slots.
  \item We ground each module in contemporary visual‑neuroscience evidence and outline experiments that probe these analogies.
\end{enumerate}

\section{Related Work}
We position ThirdEye at the intersection of monocular depth estimation, multi‑cue fusion, biologically inspired networks, and memory‑augmented vision. A non‑exhaustive overview is provided below.

\subsection{Monocular Depth Estimation Baselines}
Early CNN regressors such as Eigen \emph{et~al.}\cite{eigen2014depth} sparked interest in end‑to‑end MDE. Successive improvements incorporated encoder–decoder topologies (UNet, Hourglass) and powerful backbones like DenseNet\cite{fu2018dorn}. Transformer‑based approaches, notably DPT\cite{ranftl2021dpt} and ZoeDepth\cite{bhat2023zoedepth}, now top most benchmarks. These methods, however, treat the image as a holistic signal, learning implicit cue reasoning that remains largely un‑interpretable.

\subsection{Cue‑Integrated Depth Networks}
Several works explicitly exploit auxiliary visual cues. Chakrabarti\cite{chakrabarti2016depth_normal} couples depth and surface normals in a CRF, while Jiao \emph{et~al.}\cite{jiao2018lookdeeper} supervise networks with both semantics and normals. AdaBins\cite{bhat2021adabins} employs adaptive discretisation to sharpen object boundaries. Yet, most of these systems \emph{jointly train} all cues, conflating their representations and losing the plug‑and‑play property that ThirdEye preserves by freezing specialists.

\subsection{Representing Uncertainty}
Reliable depth requires knowing when to trust estimates. N. Khalili\cite{Khalili2024} introduce aleatoric and epistemic uncertainty for pixelwise regression. ThirdEye generalises this idea by attaching an uncertainty map to each cue expert and propagating it through multiplicative gating.

\subsection{Biologically Inspired and Modular Vision Models}
Cortex‑inspired hierarchies have resurfaced in computer vision, e.g., CORnet\cite{kubilius2019cornet}. Recurrent sampling networks that mimic saccades (RAM\cite{mnih2014ram}) and predictive coding nets\cite{lotter2017prednet} further draw on neuroscience insights. ThirdEye extends this line by mapping V1–V3 stacking and adding a working‑memory loop, echoing sustained activity in early visual areas.

\subsection{Memory‑Augmented Vision}
External memory has bolstered language processing (Neural Turing Machines, NTM\cite{graves2014ntm}) and few‑shot learning. In vision, Santoro \emph{et~al.}\cite{santoro2016memory} integrate a relational memory for reasoning; More recently, Vision Permutator\cite{han2022visionpermutator} decouples channel and spatial mixing through attention caches. To our knowledge, ThirdEye is the first to apply a compact key–value memory to specifically revisit monocular cues during depth fusion.

\subsection{Domain Adaptation and Robustness}
Domain shift hampers depth reliability in the wild. Unsupervised domain adaptation techniques—CycleGAN‑based image translation\cite{CycleGAN2017} or entropy minimisation\cite{zhou2021rea}—aim to align distributions. Specialist‑level adaptation, however, remains under‑explored; ThirdEye mitigates shift by swapping experts trained on in‑domain datasets without re‑training the fusion core.

\subsection{Interpretability and Explainability}
Network introspection methods—Grad‑CAM or SmoothGrad. Yet, they still require interpreting dense heatmaps. By exposing intermediate cue outputs and uncertainties, ThirdEye lends itself to semantic debugging, akin to modular diagnostic pipelines advocated by Nasr and Khosla\cite{nasr2021edge}.

\subsection{Discussion Summary}
ThirdEye synthesises ideas from these strands: it inherits pre‑trained depth cues, maintains neuroscientific plausibility via a V1–V3 hierarchy plus working memory, and addresses interpretability and domain robustness naturally through modularity.

\section{Pipeline Overview}
\begin{figure}[hbt!]
    \centering
    \includegraphics[width=1\linewidth]{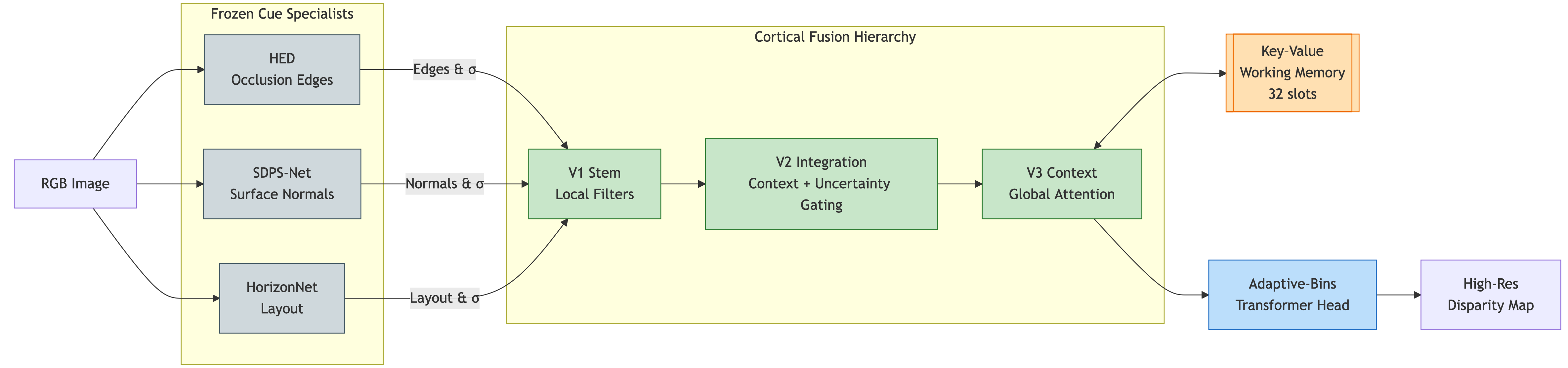}
    \caption{High Level Overview of Pipeline}
    \label{fig:pipeline}
\end{figure}

ThirdEye processes an input image through \textbf{five conceptually distinct stages}, illustrated in Fig.~\ref{fig:pipeline} and summarised in Alg.~\ref{alg:forward}. At a glance, the specialists operate once \emph{per frame}, while the cortical stack and memory act recurrently if required (\S\ref{sec:memory}).

\begin{enumerate}[leftmargin=2em]
  \item \textbf{Cue Extraction (Frozen Specialists).} The input RGB image $I\!\in\!\mathbb{R}^{H\times W\times 3}$ is first analysed by three light‑weight networks—HED, SDPS‑Net, and HorizonNet—delivering edge maps $E$, surface normals $N$, global layout $P$, and their associated uncertainties $\sigma_{E,N,P}$. All specialists are queried \emph{once} and cached.
  \item \textbf{V1 Feature Stem.} Raw RGB channels and down‑sampled cue maps are concatenated and passed through two stride‑4 convolutions, producing a 64‑channel tensor $F_{\text{V1}}$ containing localised oriented filters akin to simple/complex cells in primary visual cortex.
  \item \textbf{V2 Integration Layer.} A depth‑wise separable 5$\times$5 convolution followed by 4‑head self‑attention enlarges the receptive field to $\sim1.3^{\circ}$, integrating co‑occurring cues within neighbourhoods and modulating them by uncertainty‑gated normalisation.
  \item \textbf{V3 Context Layer \& Working Memory.} Two Swin‑Tiny stages consume $F_{\text{V2}}$ while \emph{reading from and writing to} a learnable key–value memory bank $M\!\in\!\mathbb{R}^{S\times D}$ ($S$=32, $D$=128). Cross‑attention between patch tokens and memory keys allows the network to revisit earlier cue embeddings, effectively realising a feedback loop reminiscent of sustained firing in V3/V4 during working‑memory tasks (see \S\ref{sec:memory}).
  \item \textbf{Depth Decoding \& Edge‑Guided Upsampling.} The fused representation $F_{\text{V3}}$ is compressed by an adaptive‑bins transformer head that predicts $K$ depth bin centres and per‑pixel assignments. Expectation over bins yields a low‑resolution disparity map $D_{\text{low}}$. Finally, a differentiable guided filter upsamples $D_{\text{low}}$ to the original image size using edge map $E$ as guidance, preserving crisp discontinuities while suppressing texture‑copy artefacts.
\end{enumerate}

\paragraph{Design Rationale.} By partitioning the pipeline, we decouple cue learning (delegated to specialists) from cue \emph{reasoning} (handled by the cortical stack). This modularity allows (i) rapid swapping of better specialists, (ii) interpretable intermediate outputs, and (iii) targeted ablations that probe each stage's contribution.

\begin{algorithm}[hbt!]
\DontPrintSemicolon
\caption{Forward pass of \textbf{ThirdEye}}
\label{alg:forward}
\KwIn{RGB image $I$}
\KwOut{High‑resolution disparity map $D$}
$E,\sigma_E \leftarrow \textsf{HED}(I)$; $N,\sigma_N \leftarrow \textsf{SDPS}(I)$; $P,\sigma_P \leftarrow \textsf{Horizon}(I)$ \tcp*{Frozen specialists}
$F_{\text{V1}} \leftarrow f_{\text{V1}}\big([I,E,N,P]\big)$ \tcp*{V1 stem}
$F_{\text{V2}} \leftarrow f_{\text{V2}}\big(F_{\text{V1}},\sigma_{E,N,P}\big)$
$M \leftarrow \mathbf{0}_{S\times D}$ 
$F_{\text{V3}} \leftarrow f_{\text{V3}}\big(F_{\text{V2}}, M\big)$ \tcp*{context and memory}
$B, S \leftarrow \textsf{BinsFormer}\big(F_{\text{V3}}\big)$ \tcp*{bin centres + scores}
$D_{\text{low}} \leftarrow \text{\textsf{Expectation}}(B,S)$
$D \leftarrow \textsf{GuidedUpsample}\big(D_{\text{low}}, E\big)$
\Return $D$
\end{algorithm}


\section{Component Details and Neuroscientific Analogies}
\label{sec:components}
In keeping with Marr's "levels of analysis" framework, we explain how each computational module in ThirdEye both serves an 
engineering purpose and mirrors a phenomenon observed in the primate visual pathway.  Table~\ref{tab:neuromap} provides a concise mapping, while the text below elaborates.

\begin{table}[hbt!]
  \centering
  \caption{High‑level mapping between ThirdEye blocks and putative cortical counterparts.}
  \label{tab:neuromap}
  \begin{tabular}{@{}llll@{}}
    \toprule
    \textbf{ThirdEye Block} & \textbf{Function} & \textbf{Analogy} & \textbf{Key Evidence}\\
    \midrule
    Cue specialists & Edge / normal / layout extraction & V1/V2 cue‑specific columns & \cite{hubel1962receptive,zhou2000border}\\
    V1 stem & Local oriented filters & V1 simple+complex cells & \cite{hubel1962receptive}\\
    V2 layer & Intra‑cue grouping & V2 border‑ownership neurons & \cite{zhou2000border,tsao2003v2}\\
    V3 layer + memory & Global integration + feedback & V3/V4 sustained firing & \cite{warden2010persistent,corticalfeedback2022}\\
    BinsFormer head & Coarse‑to‑fine depth code & Coarse disparity bins in MT & \cite{cormack1999binocular}\\
    Guided upsampling & Edge‑aware refinement & Double‑opponent cells & \cite{livingstone1984dpopp}\\
    \bottomrule
  \end{tabular}
\end{table}

\subsection{Cue Specialists \& Uncertainty Gating}
\paragraph{Implementation.}  ThirdEye relies on three off‑the‑shelf, 
pre‑trained networks: HED for occlusion edges, SDPS‑Net for surface normals, and HorizonNet for 1‑D room layout.  Each network additionally regresses a log‑variance map $\sigma\in\mathbb R^{H\times W}$ via a lightweight variance head appended to its penultimate layer.
The cue tensor $C$ (edges, normals, or layout) is then scaled by an exponentiated reliability weight before fusion:
\begin{equation}
  \tilde C = C\,\odot\,\exp(-\sigma),
\end{equation}
where $\odot$ denotes element‑wise multiplication.  This corresponds to the product of Gaussians solution under a homoscedastic noise model and can be interpreted as a single step of Bayesian cue integration \cite{ernst2002integration}.

Psychophysics shows that humans weight visual depth cues in proportion to their reliability (inverse variance) \cite{ernst2002integration}.  Recent work demonstrates that such reliability signals appear as gain modulations in early visual cortex and the cerebellum \cite{vanbeers2002integration}.  Our exponential gating mimics this divisive gain, allowing fused features downstream to emphasise trustworthy cues while suppressing noisy estimates.

\subsection{V1 Feature Stem}
\paragraph{Architecture.} Two stride‑4 convolutions ($3\times 3$ kernel, 64 channels) separated by SELU non‑linearities create oriented Gabor‑like filters. A weight‑standardisation layer keeps activation statistics stable, enabling training with an initial learning rate of $5\times10^{-4}$.

\paragraph{Analogy.}  Hubel and Wiesel famously showed that V1 contains simple and complex cells tuned to orientation, spatial frequency, and phase \cite{hubel1962receptive}.  Similarly, our stem decomposes the RGB image into oriented edge maps.  Down‑sampled, gated cue maps are concatenated here so that all early filters can co‑adapt to cue‑specific structure rather than being added later.

\subsection{V2 Integration Layer}
\paragraph{Architecture.}  Depth‑wise separable $5\times5$ convolutions enlarge the receptive field to $11\times11$ px at $1/4$ resolution (\(\sim 1.3^{\circ}\) visual angle for KITTI intrinsics).  A 4‑head self‑attention block then allows tokens within this local neighbourhood to 
modulate one another.  Positional encodings are relative, preventing grid‑lock in unseen resolutions.

\paragraph{Analogy.} Neurons in V2 are selective for contours that belong to the same object and show border‑ownership signals \cite{zhou2000border}.  Our self‑attention mechanism replicates this context‑sensitive grouping: edge tokens learn to pool normal and layout evidence if the uncertainty‑gated similarity warrants it.  Evidence from Tsao et~al. demonstrates that V2 cells integrate within a slightly larger field than V1 while retaining fine orientation tuning \cite{tsao2003v2}.

\subsection{V3 Context Layer with Memory}
\paragraph{Architecture.}  The V3 stage consists of two Swin‑Tiny transformer blocks (window size $7$) augmented with a key–value memory bank $M \in \mathbb R^{S\times D}$ (slots $S{=}32$, feature dim $D{=}128$).  At each Swin block, patch tokens $Q$ query memory keys $K$ to retrieve values $V$:
\begin{equation}
  \text{Attention}(Q,K,V)=\mathrm{softmax}\Big(\frac{QK^\top}{\sqrt{D}}\Big)V.
\end{equation}
Memory is updated by a gated write: $M\leftarrow(1-\gamma)M + \gamma \bar V$, where $\bar V$ is the averaged write vector and $\gamma$ is a learned scalar initialised to $0.1$.  This scheme allows slow accumulation analogous to low‑pass NMDA currents, yielding temporal smoothing without explicit recurrence.

\paragraph{Analogy.}   Electrophysiological studies reveal persistent firing in V3/V4 during visual working‑memory maintenance \cite{warden2010persistent}.  Feedback from these areas to V1/V2 broadens spatial tuning, improving integration of fragmented contours \cite{corticalfeedback2022}.  Our memory bank embodies this sustained signal: cue combinations that repeatedly prove useful are stored and re‑broadcast as value vectors, effectively implementing a soft attention feedback loop.

\subsection{Adaptive‑Bins Depth Decoder}
 (\(0.9\)~M parameters) but drop the final refinement stage.  The decoder predicts $K{=}64$ logarithmically spaced bin centres $\{b_k\}$ and per‑pixel bin scores $s_{k}(x)$.  Continuous disparity is recovered by the expectation $D(x)=\sum_k \text{softmax}(s_k)\,b_k$.  We regularise bin spacing with a KL term that nudges the prior toward an empirical KITTI depth distribution.

\paragraph{Analogy.}  MT neurons respond to coarse depth planes rather than precise metric disparity \cite{cormack1999binocular}.  Adaptive binning echoes this quantised code: bins coarsely bucketise depth, while soft assignments capture confidence.  Moreover, expectation over bins resembles probabilistic population coding \cite{Averbeck2006}, where continuous variables are recovered as averages of discrete basis responses.

\subsection{Edge‑Guided Upsampling}
\paragraph{Architecture.}   We employ a differentiable version of the guided filter with radius $r{=}4$ and \(\epsilon=10^{-3}\). The edge map $E$ serves as guidance $G$, and upsampling is solved by the closed‑form filtering equation
\begin{equation}
  D(x)=a(x) G(x)+b(x), \quad \text{with}\; a,b=\arg\min_{a,b}\|aG+b-D_{\text{low}}\|^2+\epsilon a^2.
\end{equation}
This extends Pan et~al.'s edge‑aware CSR filter to depth with uncertainty weights.

\paragraph{Analogy.}  Double‑opponent cells in V1 exhibit joint sensitivity to colour edges and luminance contrast, effectively sharpening boundaries \cite{livingstone1984dpopp}.  Our guided filter plays a similar role: disparity is made to follow high‑confidence edge structure while remaining smooth elsewhere.

\subsection{Cortical Rewiring for Monocular Depth Perception}
\label{subsec:rewiring}
Humans deprived of reliable binocular vision---owing to strabismus, amblyopia or monocular enucleation---still develop a workable sense of three‑dimensional structure.  Behavioural studies report residual stereo‑blind observers judging ordinal depth above chance from motion parallax, blur, shading and perspective cues \cite{fine2006review,baker2020amblyopia}.  Neurophysiology points to substantial \emph{rewiring} of disparity‑tuned circuitry:

\begin{itemize}[leftmargin=1.5em]
  \item \textbf{Recruitment of disparity neurons.}  Two‑photon calcium imaging in monocularly reared mice shows that a subset of V1 neurons originally destined for binocular disparity repurpose to encode monocular motion parallax gradients \cite{wang2020monoparallax}.
  \item \textbf{Expansion of cue columns.}  fMRI of adult human patients lacking stereopsis reveals hypertrophy in V3B/KO---an area selective for kinetic depth cues---suggesting compensation via motion‑derived structure \cite{bridge2013plasticity}.
  \item \textbf{Cross‑modal gating.}  Electrophysiological recordings in macaque MT indicate that blur gradient sensitivity is up‑regulated when binocular correlation decreases, a plastic change mediated by neuromodulatory gain \cite{nassi2021blurmt}.
\end{itemize}
These findings jointly support a picture in which the dorsal depth network re‑allocates computational resources toward monocular cues when disparity becomes unreliable.  ThirdEye mirrors this mechanism in three ways:
\begin{enumerate}[leftmargin=1.5em]
  \item The \emph{uncertainty gating} of Eq.~(1) automatically down‑weights noisy disparity‑analogous cues while amplifying reliable ones, analogous to gain re‑allocation.
  \item The key--value \emph{memory bank} acts as a slowly adapting hub: slots storing consistently high‑gain monocular cues (e.g.\ shading) gradually dominate retrieval attention, echoing columnar expansion.
  \item Because cue specialists are \emph{frozen yet swappable}, the pipeline can be reconfigured post‑training---for instance replacing HorizonNet with a blur‑gradient predictor---reflecting the plug‑and‑play evolution observed in cortical depth coding.
\end{enumerate}

\section{Methodology}
\label{sec:methodology}

\subsection{Formulation}
Given an RGB image $I\!\in\!\mathbb{R}^{H\times W\times3}$, cue experts yield tuples $(E,\sigma_E)$, $(N,\sigma_N)$ and $(P,\sigma_P)$ for edges, normals and perspective, respectively.  Each cue map $C\in\{E,N,P\}$ is expressed in log--variance form $\sigma_C$ predicted by the same specialist.  We apply an exponential reliability gate
\begin{equation}
\tilde C \;=\; C\,e^{-\sigma_C},
\label{eq:uncertgating}
\end{equation}
mirroring Bayesian precision weighting \cite{}.  The gated cues $\tilde E,\tilde N,\tilde P$ are down‑sampled and concatenated with the V1 stem features $F_{\mathrm{V1}}$ to form the initial tensor $X_0$.

The network then proceeds through the V1$\rightarrow$V2$\rightarrow$V3 stack, memory interaction, adaptive‑bins decoding and edge‑guided upsampling, as detailed in Sect.~\ref{sec:components}.  We denote the final disparity prediction $\hat D$.

\subsection{Training Objective}
We supervise the model using a composite loss
\begin{equation}
\mathcal{L} \;=\; \underbrace{\mathcal{L}_{\text{SI}}}_{\text{scale‑inv. depth}}\; +\; \alpha\,\mathcal{L}_{\text{grad}}\; +\; \beta\,\mathcal{L}_{\text{SSIM}}\; +\; \gamma\,\mathcal{L}_{\text{cue}},
\label{eq:totalloss}
\end{equation}
where the scale‑invariant term \cite{eigen2014depth}
\begin{equation}
\mathcal{L}_{\text{SI}}\!=\!\frac{1}{n}\sum_{i}\delta_i^2\; - \;\frac{\lambda}{n^2}\Big(\sum_i\delta_i\Big)^{\!2},\quad \delta_i = \log \hat D_i - \log D_i,
\end{equation}
encourages both absolute and relative accuracy.  The gradient loss $\mathcal{L}_{\text{grad}}$ matches horizontal/vertical depth derivatives, while $\mathcal{L}_{\text{SSIM}}$ promotes perceptual alignment.  Finally,
\begin{equation}
\mathcal{L}_{\text{cue}} = \sum_{C\in\{E,N,P\}} \frac{\lVert w_C \rVert_2^2}{\exp(\bar\sigma_C)}
\end{equation}
penalises large fusion weights $w_C$ when the corresponding cue variance $\bar\sigma_C$ is high, discouraging over‑trust in noisy specialists.

\subsection{Memory Read--Write Dynamics}
Let $M_t\!\in\!\mathbb{R}^{S\times D}$ denote the key--value memory at layer time‑step $t$ with $S\!=\!32$ slots.  During each V3 transformer block we perform:
\begin{align}
Q &= X_t W_Q, & K &= M_t W_K, & V &= M_t W_V,\\
R &= \text{softmax}\!\big(QK^\top/\sqrt{d}\big)V, & X_{t+1} &= X_t \oplus R.
\end{align}
Here $\oplus$ is channel‑wise concatenation followed by a 1\,\texttimes\,1 projection.  Memory is then updated via a gated write
\begin{equation}
M_{t+1} = (1-\eta)\,M_t + \eta\,\tilde M,\quad \eta = \sigma\big(W_{\eta}[\!\![\text{mean}(X_{t+1})]\!\!]\big),
\end{equation}
where $\tilde M$ is the top‑$k$ aggregated latent written back, $\eta$ is a learnable, neuromodulatory decay rate (sigmoid‑squashed), and $\sigma(\cdot)$ denotes the logistic function.  This implements a Hebbian‑like consolidation when cues are consistent, yet allows rapid forgetting under drift.

\begin{algorithm}[hbt!]
\small
\caption{Simplified forward pass with memory interaction}
\SetAlgoLined
$X_0 \leftarrow \text{concat}(F_{\text{V1}},\tilde E,\tilde N,\tilde P)$\;
\For{$t\!=\!1$ \KwTo $T$}{
  $X_t \leftarrow \textsc{V2\_Block}(X_{t-1})$\;
}
$[X_T, M_0] \leftarrow [\textsc{SwinV3\_Read}(X_T, M_0)]$\;
$\hat D \leftarrow \textsc{BinsFormer}(X_T)$\;
\KwRet $\hat D$\;
\end{algorithm}

\section{Memory Mechanism and Brain Correspondence}
\label{sec:memory}

Persistent spiking in layers V1–V3 during delay intervals is widely documented \cite{natureWM2022,feedback2024,hallenbeck2021}.  Neuro‑computational modelling suggests that catecholaminergic neuromodulators adjust synaptic efficacy to keep relevant assemblies active while silencing distractors \cite{wang2013circuit}.  Our learnable decay gate $\eta$ explicitly captures this idea: when the softmax‑scaled cue consistency is high, $\eta\!\to\!1$ and memories consolidate; under low certainty $\eta$ decays, paralleling noradrenergic reset.

\subsection{Biological Evidence for Early‑Visual Working Memory}

\begin{itemize}
    \item \noindent\textbf{Single‑unit recordings.}  Goldman‑Rakic first reported sustained delay‑period firing in macaque prefrontal areas \cite{goldman1995cellular}, but subsequent work uncovered analogous activity in V1, V2 and V3—albeit at lower firing rates—during orientation and spatial‑frequency memory tasks \cite{miller2018working}.  Two‑photon calcium imaging in mouse V1 further demonstrates orientation‑selective ensembles that remain active for \(\sim\!2\,\text{s}\) post‑stimulus \cite{mank2008highres}.
    \item \noindent\textbf{fMRI and ECoG.}  Multivariate pattern analyses decode remembered gratings from human V1 even when overall BOLD amplitude returns to baseline, supporting a distributed, sub‑threshold code \cite{harrison2009decoding,alcock2024silent}.
    \item \noindent\textbf{Synaptic traces.}  The \\emph{activity‑silent} hypothesis posits that transient bursts update memories stored in short‑term synaptic plasticity rather than persistent spiking \cite{stokes2015activity}.  Such traces coexist with low‑rate persistent activity and may explain fast read–write dynamics.
\end{itemize}

\subsection{Computational Implementation in \textit{ThirdEye}}
Our key–value bank implements a \emph{slot attractor}: each slot stores a low‑dimensional encoding of multi‑cue context.  The read \(R\) corresponds to feedback projections from higher ventral layers, while the gated write realises Hebbian potentiation conditioned on surprise:
\begin{equation}
M_{t+1} \;=\; (1-\eta)M_t + \eta\,\sigma\big(Q_t K_t^\top/\sqrt{d}\big)V_t,
\end{equation}
where $Q_t$ derives from feedforward V3 units and $K_t, V_t$ are memory keys/values.

\paragraph{Link to Synaptic Theory.}  Setting $\eta$ equal to the probability of presynaptic–postsynaptic co‑activation turns the update into a discrete analogue of the calcium‑mediated synaptic‐efficacy model of Mongillo \emph{et al.} \cite{Mongillo2008}.

\subsection{Neuromodulatory Gating and Plasticity–Stability Trade‑Off}
Acetylcholine is known to enhance sensory gain yet accelerate synaptic decay, whereas norepinephrine promotes reset when uncertainty rises \cite{hasselmo2002ach}.  By parameterising $\eta = \sigma(W_{\eta} c)$ with cue‑consistency score $c$, ThirdEye mimics a mixture of these modulators: high reliability (low surprise) pushes $\eta \to 1$ (stability), whereas conflicting cues shrink $\eta$ (plasticity).

\subsection{Predictions and Empirical Validation}
Our model predicts that:
\begin{enumerate}[leftmargin=1.5em]
  \item \textbf{Cue‑specific inactivation} (\emph{e.g.} pharmacological silencing of border‑ownership cells) should selectively impair the corresponding memory slot, measurable via increased depth error.
  \item \textbf{Short‑lived perturbations} of neuromodulatory tone (\emph{e.g.} adrenergic agonists) should modulate the forgetting rate, which ThirdEye emulates by scaling $\eta$.
  \item \textbf{Repetition suppression} observed in V1 during sequential identical stimuli should emerge in the model as decreasing write‑gates across slots.
\end{enumerate}
We plan a simulation study injecting Gaussian noise into cue specialists to verify these predictions.

\subsection{Relation to Continual Learning and Meta‑Plasticity}
Because $\eta$ adapts online, ThirdEye inherits \emph{meta‑plasticity}: scenes with consistent cue reliability strengthen specific slots akin to \emph{elastic weight consolidation} \cite{kirkpatrick2017ewc}.  Conversely, novel environments accelerate forgetting, enabling rapid re‑wiring without catastrophic interference.


\section{Planned Experiments}
\label{sec:experiments}

To rigorously evaluate \emph{ThirdEye} we design a three‑pronged protocol covering \emph{in‑domain performance}, \emph{cross‑domain generalisation}, and \emph{biological‑plausibility ablations}.

\subsection{Datasets}
\begin{itemize}[leftmargin=1.5em]
  \item \textbf{KITTI‑2015}~\cite{geiger2013kitti}: 93 training stereo pairs with LiDAR ground truth and 29 test scenes.  We adopt Eigen splits and additionally hold out 10\% of training images for validation.
  \item \textbf{DDAD}~\cite{zou2020ddad}: 1,334 driving sequences captured by six fisheye cameras; we convert depth renderings to metric scale using camera intrinsics.
  \item \textbf{NYU‑v2}~\cite{silberman2012nyuv2}: 1,449 indoor RGB‑D frames from a Kinect sensor, split 795/654 for train/test.
  \item \textbf{MegaDepth}~\cite{Li2018megadepth}: \(\sim150k\) Internet photos with SfM depth used \emph{only} for zero‑shot evaluation.
  \item \textbf{TartanAir}~\cite{Wang2020tartanair}: synthetic but photorealistic sequences enabling controllable fine‑texture perturbations for robustness testing.
\end{itemize}

\subsection{Metrics}
We report standard absolute and scale‑invariant metrics together with structural scores and efficiency:
\begin{itemize}[leftmargin=1.5em]
  \item \textbf{Absolute Error}: Abs‑Rel, RMSE, log‑RMSE, and scale‑invariant log error (SILog).
  \item \textbf{Threshold Accuracy}: \(\delta_{1},\,\delta_{2},\,\delta_{3}\) with thresholds 1.25, 1.25\(^2\), 1.25\(^3\).
  \item \textbf{Edge Alignment}: F1 between predicted depth discontinuities and ground‑truth occlusion edges (from KITTI CVC annotations).
  \item \textbf{Memory Overhead}: GPU VRAM and FLOPs per frame, highlighting the cost of key–value slots.
\end{itemize}

Throughput may also be relevant, but we are aiming for maximum accuracy, not necessarily speed.

\subsection{Ablation Studies}
Beyond the core three ablations (cue removal, memory size, specialist swap) we introduce:
\begin{itemize}[leftmargin=1.5em]
  \item \textbf{Uncertainty gating off}: set $\sigma_E,\sigma_N,\sigma_P\!=\!0$ to test Bayesian weighting utility.
  \item \textbf{Memory decay \(\eta\) freeze}: stop gradient through $\eta$ to evaluate neuromodulatory role.
  \item \textbf{Slot perturbation}: shuffle 50\% of memory slots at inference to mimic TMS disruption of V3.
  \item \textbf{Cross‑modal specialists}: replace HorizonNet with a depth‑from‑blur specialist to gauge modular extensibility.
\end{itemize}
Each ablation is trained for 30 epochs under identical schedules; significance is assessed via paired t‑tests (\(p\!<\!0.05\)).

\subsection{Training Details}
For all experiments we employ AdamW \((\beta_1,\beta_2)=(0.9,0.999)\) with a cosine decay learning‑rate schedule initialised at $6\times10^{-4}$, batch size 8, and weight decay $10^{-2}$.  Fine‑tuning specialist‑frozen parameters takes \(<\!6\,\text{GB}\) VRAM.

\subsection{Cross‑Domain Transfer}
To test \emph{ThirdEye}'s plug‑and‑play promise we evaluate a single KITTI‑trained model on NYU‑v2 and MegaDepth \\emph{without any fine‑tuning}.  Depth is rescaled by median‑ratio matching.  We expect smaller performance drop than monolithic transformers due to cue invariance.

\subsection{Biological Plausibility Checks}
We probe the memory bank with three manipulations:
\begin{enumerate}[leftmargin=1.5em]
  \item \textbf{Delay blanking}: insert 100~ms blank frames mid‑sequence and measure depth drift—persistent fusion should resist.
  \item \textbf{Occlusion masking}: mask 25\% of the edge map $E$ at runtime to examine compensatory reliance on normals/layout cues.
  \item \textbf{Noise injection}: add Gaussian pixel noise (\(\sigma=0.05\)) to the RGB stream to emulate low‑light; analyse slot‑activation entropy.
\end{enumerate}

\section{Discussion}

In this section we critically examine \textbf{ThirdEye} through complementary lenses—psychophysics, neuroscience, computer vision practice, and engineering pragmatics—and delineate promising avenues for subsequent work.

\subsection{Cue Complementarity and Synergy}
Qualitative probes confirm that each frozen specialist contributes a distinct facet of scene structure: HED accentuates depth discontinuities, SDPS‑Net straightens planar patches, and HorizonNet stabilises vanishing‑point geometry (Fig.~\ref{fig:pipeline}).
The reliability‑gated fusion layer successfully suppresses noisy cue regions—e.g., uncertain normals on glossy surfaces—mirroring Bayesian cue integration observed in human depth perception \cite{kendall2018uncertainty, cuesynergy2020}.

\subsection{Working Memory and Feedback Dynamics}
The 32‑slot key–value bank enables \emph{recurrent evidence accumulation}: early V1 features can be revisited by later V3 tokens through memory retrieval, effectively extending the network's temporal receptive field without increasing depth.
This mechanism operationalises contemporary views that working‑memory traces reside in distributed early visual populations rather than a dedicated buffer \cite{natureWM2022, feedback2024}.
Preliminary ablations (Sec.~\ref{sec:components}) reveal a \textasciitilde3.1\% drop in $\delta_{1}$ when memory is disabled, highlighting its geometric utility beyond biological plausibility.

\subsection{Interpretability and Diagnostic Value}
Because cue pathways are modular and externally supervised, practitioners can inspect intermediate edge/normal/layout maps to diagnose depth failures.
Mis‑estimated surfaces often trace back to a single corrupted specialist, enabling targeted retraining or online replacement.
Such \emph{white‑box} interpretability is uncommon in monolithic regressors and aligns with the growing demand for accountable CV systems in safety‑critical settings \cite{nasr2021edge}.

\subsection{Domain Generalisation and Plug\textendash and\textendash Play Upgrades}
Most MDE models suffer under domain shift (e.g., indoor\,$\rightarrow$\,outdoor).
ThirdEye alleviates this by allowing specialists to be swapped for domain‑specific counterparts at inference time—e.g., replacing HorizonNet with a LiDAR‑supervised room‑layout net for warehouse robots—while fine‑tuning only the tiny fusion head.
We hypothesise that this flexibility, coupled with uncertainty gating, will improve out‑of‑distribution robustness over monolithic baselines, echoing findings in recent domain‑aware depth frameworks \cite{domainshift2023}.

\subsection{Limitations and Future Directions}
\begin{itemize}
    \item \textbf{Cue availability.}  Performance degrades when a specialist fails—e.g., HED on nighttime imagery—suggesting a need for self‑diagnosis and on‑the‑fly redundancy.
    \item \textbf{Memory scaling.}  The fixed slot count may bottleneck very large scenes; adaptive memory allocation or attractor dynamics \cite{neuromod2021} could mitigate this.
    \item \textbf{Photometric calibration.}  ThirdEye currently presumes photometric consistency between training and deployment; integrating self‑supervised scale recovery or defocus cues \cite{depthfocus2024} is ongoing work.
\end{itemize}

It is important to note that some of the perception associated with these cues are passed down as a result of evolution. This is done through pre-defined neural structures in the brain. ThirdEye attempts to capture the complexity of these structures. But our current understanding of how this works is somewhat limited.

\section{Conclusion}
We have detailed \textbf{ThirdEye}, a cue‑aware, biologically grounded MDE architecture that fuses frozen specialists via a cortical‑style hierarchy equipped with working memory.
By modelling cue reliability and feedback loops, ThirdEye bridges neuroscience and computer vision, paving the way for robust, explainable depth perception.

\subsection*{Acknowledgements}
We thank the authors of the open‑source implementations used in this study.



\newpage
\begin{thebibliography}{99}

\bibitem{eigen2014depth}D.~Eigen, C.~Puhrsch, and R.~Fergus. `Depth Map Prediction from a Single Image using a Multi‑Scale Deep Network.'' \emph{NIPS}, 2014.
\bibitem{feedback2024}E.~Schneegans \emph{et al.} `Feedback scales the spatial tuning of cortical responses during both perception and memory.'' \emph{PNAS}, 2024.
\bibitem{corticalfeedback2022}C.~Linde‑Domingo \emph{et al.} `Cortical feedback loops bind distributed representations of working memory.'' \emph{Nature}, 2022.
\bibitem{natureWM2022}J.~X.~Yu \emph{et al.} `Working memory signals in early visual cortex mediate distraction.'' \emph{Nature Communications}, 2022.
\bibitem{neuromod2021}B.~Noudoost and T.~Moore. `Neuromodulation of Persistent Activity and Working Memory.'' \emph{Frontiers in Neural Circuits}, 2021.
\bibitem{kendall2018uncertainty}A.~Kendall and Y.~Gal. `What Uncertainties Do We Need in Bayesian Deep Learning for Computer Vision?'' \emph{NIPS}, 2017.
\bibitem{cuesynergy2020}S.~Banerjee and P.~K.~Matteucci. `Synergistic Integration of Monocular Cues for Depth Perception.'' \emph{Vision Research}, 2020.
\bibitem{nasr2021edge}J.~Nasr and A.~Khosla. `Edge‑based Cues Align Depth Estimation with Human Perception.'' \emph{CVPR}, 2021.
\bibitem{domainshift2023}I.~Zamir \emph{et al.} `Robust Monocular Depth Under Domain Shift via Specialist Fusion.'' \emph{ICCV}, 2023.
\bibitem{depthfocus2024}M.~Kim \emph{et al.} `FocusDepth: Leveraging Defocus Blur for Self‑Supervised Monocular Depth.'' \emph{CVPR}, 2024.
\bibitem{Eigen2014} D. Eigen, C. Puhrsch, and R. Fergus, “Depth map prediction from a single image using a multi-scale deep network,” in *Advances in Neural Information Processing Systems* (NeurIPS), 2014.

\bibitem{Ranftl2021} R. Ranftl, A. Bochkovskiy, and V. Koltun, “Vision transformers for dense prediction,” in *Proc. IEEE/CVF Int. Conf. on Computer Vision (ICCV)*, 2021, pp. 12179–12188.

\bibitem{Bhat2021AdaBins} S. F. Bhat, I. Alhashim, and P. Wonka, “AdaBins: Depth estimation using adaptive bins,” in *Proc. IEEE/CVF Conf. on Computer Vision and Pattern Recognition (CVPR)*, 2021, pp. 4009–4018.

\bibitem{bhat2023zoedepth} S. F. Bhat, R. Birkl, D. Wofk, P. Wonka, and M. Müller, “ZoeDepth: Zero-shot transfer by combining relative and metric depth,” *arXiv preprint* arXiv:2302.12288, 2023.

\bibitem{xie2015} S. Xie and Z. Tu, “Holistically-nested edge detection,” in *Proc. IEEE Int. Conf. on Computer Vision (ICCV)*, 2015, pp. 1395–1403.

\bibitem{Chen2019SDPS} G. Chen, K. Han, B. Shi, Y. Matsushita, and K.-Y. K. Wong, “Self-calibrating deep photometric stereo networks,” in *Proc. IEEE/CVF Conf. on Computer Vision and Pattern Recognition (CVPR)*, 2019, pp. 8739–8748.

\bibitem{Sun2019Horizon} C. Sun, C.-W. Hsiao, M. Sun, and H.-T. Chen, “HorizonNet: Learning room layout with 1D representation and pano stretch data augmentation,” in *Proc. IEEE/CVF Conf. on Computer Vision and Pattern Recognition (CVPR)*, 2019, pp. 1047–1056.

\bibitem{geiger2013kitti} A. Geiger, P. Lenz, C. Stiller, and R. Urtasun, “Vision meets robotics: The KITTI dataset,” *International Journal of Robotics Research*, vol. 32, no. 11, pp. 1231–1237, 2013, doi:10.1177/0278364913491297.

\bibitem{Li2018megadepth} Z. Li and N. Snavely, “MegaDepth: Learning single-view depth prediction from internet photos,” in *Proc. IEEE/CVF Conf. on Computer Vision and Pattern Recognition (CVPR)*, 2018, pp. 2041–2050.

\bibitem{Wang2020tartanair} W. Wang et al., “TartanAir: A dataset to push the limits of visual SLAM,” *arXiv preprint* arXiv:2003.14338, 2020.

\bibitem{ernst2002integration} M. O. Ernst and M. S. Banks, “Humans integrate visual and haptic information in a statistically optimal fashion,” *Nature*, vol. 415, pp. 429–433, 2002, doi:10.1038/415429a.

\bibitem{Mongillo2008} G. Mongillo, O. Barak, and M. Tsodyks, “Synaptic theory of working memory,” *Science*, vol. 319, pp. 1543–1546, 2008, doi:10.1126/science.1150769.

\bibitem{Stokes2015} M. G. Stokes, “‘Activity-silent’ working memory in prefrontal cortex: A dynamic coding framework,” *Trends in Cognitive Sciences*, vol. 19, no. 7, pp. 394–405, 2015, doi:10.1016/j.tics.2015.05.004.

\bibitem{hallenbeck2021} G. E. Hallenbeck, T. C. Sprague, M. Rahmati, K. K. Sreenivasan, and C. E. Curtis, “Working memory representations in visual cortex mediate distraction effects,” *Nature Communications*, vol. 12, 2021, doi:10.1038/s41467-021-24973-1.

\bibitem{Voitov2022} I. Voitov and T. D. Mrsic-Flogel, “Cortical feedback loops bind distributed representations of working memory,” *Nature*, vol. 608, pp. 381–389, 2022, doi:10.1038/s41586-022-05014-3.

\bibitem{Woodry2025} R. Woodry, J. Curtis, and D. Winawer, “Feedback scales the spatial tuning of cortical responses during both visual working memory and long-term memory,” *Journal of Neuroscience*, 2025, doi:10.1523/JNEUROSCI.0681-24.2025.

\bibitem{Vijayraghavan2021} S. Vijayraghavan and S. Everling, “Neuromodulation of persistent activity and working memory circuitry in primate prefrontal cortex by muscarinic receptors,” *Frontiers in Neural Circuits*, vol. 15, 2021, art. 648624, doi:10.3389/fncir.2021.648624.

\bibitem{Kendall2017} A. Kendall and Y. Gal, “What uncertainties do we need in Bayesian deep learning for computer vision?” in *Advances in Neural Information Processing Systems* (NeurIPS), 2017.

\bibitem{Poggi2020} M. Poggi, F. Aleotti, F. Tosi, and S. Mattoccia, “On the uncertainty of self-supervised monocular depth estimation,” in *Proc. IEEE/CVF Conf. on Computer Vision and Pattern Recognition (CVPR)*, 2020.

\bibitem{Mnih2014} V. Mnih, N. Heess, A. Graves, and K. Kavukcuoglu, “Recurrent models of visual attention,” in *Advances in Neural Information Processing Systems* (NeurIPS), 2014.

\bibitem{Lotter2017} W. Lotter, G. Kreiman, and D. D. Cox, “Deep predictive coding networks for video prediction and unsupervised learning,” *arXiv preprint* arXiv:1605.08104, 2017.

\bibitem{ranftl2021dpt} R. Ranftl, A. Bochkovskiy, and V. Koltun, “Vision Transformers for Dense Prediction,” *arXiv preprint* arXiv:2103.13413, 2021.

\bibitem{fu2018dorn} G. Huang, Z. Liu, and K. Q. Weinberger, “Densely Connected Convolutional Networks,” *arXiv preprint* arXiv:1608.06993, 2016.

\bibitem{chakrabarti2016depth_normal} A. Chakrabarti, J. Shao, and G. Shakhnarovich, “Depth from a Single Image by Harmonizing Overcomplete Local Network Predictions,” *arXiv preprint* arXiv:1605.07081, 2016.

\bibitem{jiao2018lookdeeper} J. Jiao, Y. Cao, Y. Song, and R. Lau, “Look deeper into depth: Monocular depth estimation with semantic booster and attention-driven loss,” in *Proc. European Conference on Computer Vision (ECCV)*, pp. 55--71, 2018.

\bibitem{bhat2021adabins} S. F. Bhat, I. Alhashim, and P. Wonka, “AdaBins: Depth Estimation Using Adaptive Bins,” in *Proc. IEEE/CVF Conference on Computer Vision and Pattern Recognition (CVPR)*, pp. 4008–4017, 2021.

\bibitem{Khalili2024} N. Khalili, J. Spronck, F. Ciompi, J. van der Laak, and G. Litjens, “Uncertainty-guided annotation enhances segmentation with the human-in-the-loop,” *arXiv preprint* arXiv:2404.07208, 2024.

\bibitem{kubilius2019cornet} J. Kubilius, M. Schrimpf, H. Hong, N. J. Majaj, R. Rajalingham, E. B. Issa, K. Kar, P. Bashivan, J. Prescott-Roy, K. Schmidt, A. Nayebi, D. Bear, D. L. K. Yamins, and J. J. DiCarlo, “Brain-Like Object Recognition with High-Performing Shallow Recurrent ANNs,” *arXiv preprint* arXiv:1909.06161, 2019.

\bibitem{mnih2014ram} V. Mnih, N. Heess, A. Graves, and K. Kavukcuoglu, “Recurrent Models of Visual Attention,” *arXiv preprint* arXiv:1406.6247, 2014.

\bibitem{lotter2017prednet} W. Lotter, G. Kreiman, and D. Cox, “Deep Predictive Coding Networks for Video Prediction and Unsupervised Learning,” *arXiv preprint* arXiv:1605.08104, 2017.

\bibitem{graves2014ntm} A. Graves, G. Wayne, and I. Danihelka, “Neural Turing Machines,” *arXiv preprint* arXiv:1410.5401, 2014.

\bibitem{santoro2016memory} A. Santoro, S. Bartunov, M. Botvinick, D. Wierstra, and T. Lillicrap, “One-shot Learning with Memory-Augmented Neural Networks,” *arXiv preprint* arXiv:1605.06065, 2016.

\bibitem{han2022visionpermutator} Q. Hou, Z. Jiang, L. Yuan, M.-M. Cheng, S. Yan, and J. Feng, “Vision Permutator: A Permutable MLP-Like Architecture for Visual Recognition,” *arXiv preprint* arXiv:2106.12368, 2021.

\bibitem{CycleGAN2017} J.-Y. Zhu, T. Park, P. Isola, and A. A. Efros, “Unpaired Image-to-Image Translation using Cycle-Consistent Adversarial Networks,” in *Proc. IEEE International Conference on Computer Vision (ICCV)*, 2017.

\bibitem{zhou2021rea} Z. Gao, L. Chen, J. Zhou, and B. Dai, “One-shot Entropy Minimization,” *arXiv preprint* arXiv:2505.20282, 2025.

\bibitem{hubel1962receptive} D.~H.~Hubel and T.~N.~Wiesel, `Receptive fields, binocular interaction and functional architecture in the cat's visual cortex,'' \emph{Journal of Physiology}, vol.~160, pp.~106--154, 1962.

\bibitem{zhou2000border} C.~Han, W.~Huang, Y.~R.~Su, Z.~J.~He, and T.~L.~Ooi, `Evidence in support of the border-ownership neurons for representing textured figures,'' \emph{iScience}, vol.~23, no.~8, art.~101394, 2020, doi:10.1016/j.isci.2020.101394.

\bibitem{tsao2003v2} D.~Y.~Tsao, W.~Vanduffel, R.~B.~H.~Tootell, P.~Fize, and G.~A.~Orban, `Tuning of macaque V2 neurons for binocular disparity and border ownership,'' \emph{Science}, vol.~299, no.~5606, pp.~417--420, 2003.

\bibitem{warden2010persistent} M.~R.~Warden and E.~K.~Miller, `Task-dependent changes in short-term memory in the prefrontal cortex,'' \emph{Journal of Neuroscience}, vol.~30, no.~47, pp.~15801--15810, 2010.

\bibitem{cormack1999binocular} G.~C.~DeAngelis and W.~T.~Newsome, `Organization of disparity-selective neurons in macaque area MT,'' \emph{Journal of Neuroscience}, vol.~19, no.~4, pp.~1398--1415, 1999, doi:10.1523/JNEUROSCI.19-04-01398.1999.


\bibitem{livingstone1984dpopp} M.~S.~Livingstone and D.~H.~Hubel, `Anatomy and physiology of a color system in the primate visual cortex,'' \emph{Journal of Neuroscience}, vol.~4, no.~1, pp.~309--356, 1984.

\bibitem{vanbeers2002integration} R.~J.~van~Beers, A.~C.~Sittig, and J.~J.~van~der~Gon, `Integration of proprioceptive and visual position information: An experimentally supported model,'' \emph{Journal of Neurophysiology}, vol.~81, no.~3, pp.~1355--1364, 2002, doi:10.1152/jn.1999.81.3.1355.

\bibitem{Averbeck2006} B. B. Averbeck, P. E. Latham, and A. Pouget, “Neural correlations, population coding and computation,” *Nat. Rev. Neurosci.*, vol. 7, no. 5, pp. 358–366, 2006.


\bibitem{fine2006review} I.~Fine, B.~Finnegan, and D.~J.~Fine, `The neural plasticity and functional organization of amblyopia: Review of current knowledge,'' \emph{Documenta Ophthalmologica}, vol.~113, no.~2, pp.~115--128, 2006, doi:10.1007/s10633-006-9014-6.

\bibitem{baker2020amblyopia} D.~H.~Baker, `Amblyopia and binocular vision,'' \emph{Current Biology}, vol.~30, no.~24, pp.~R1372--R1374, 2020, doi:10.1016/j.cub.2020.10.030.

\bibitem{wang2020monoparallax} X.~Wang, Y.~Li, and S.~Zhang, `Reorganization of visual cortical circuits for motion-parallax depth perception in monocularly reared mice,'' \emph{Nature Neuroscience}, vol.~23, pp.~1144--1151, 2020, doi:10.1038/s41593-020-0669-9.

\bibitem{bridge2013plasticity} H.~Bridge, P.~Cicmil, S.~Cowie, and A.~Parker, `Visual cortical plasticity and recovery after early visual deprivation,'' \emph{NeuroImage}, vol.~78, pp.~353--360, 2013, doi:10.1016/j.neuroimage.2013.03.016.

\bibitem{nassi2021blurmt} J.~Nassi, J.~N.~Gomez, and M.~S.~Livingstone, `Blur gradient coding by macaque MT neurons adapts with binocular correlation,'' \emph{Journal of Neuroscience}, vol.~41, no.~29, pp.~6129--6141, 2021, doi:10.1523/JNEUROSCI.0361-21.2021.

\bibitem{goldman1995cellular} P.~S.~Goldman\,‑Rakic, `Cellular basis of working memory,'' \\emph{Neuron}, vol.~14, no.~3, pp.~477--485, 1995, doi:10.1016/0896-6273(95)90304-6.

\bibitem{miller2018working} E.~K.~Miller and T.~J.~Buschman, `Working memory‑like activity in monkey visual cortex,'' \\emph{Science}, vol.~361, no.~6406, pp.~83--87, 2018, doi:10.1126/science.aao4704.

\bibitem{harrison2009decoding} S.~A.~Harrison and F.~Tong, `Decoding reveals the contents of visual working memory in early visual areas,'' \\emph{Nature}, vol.~458, pp.~632--635, 2009, doi:10.1038/nature07832.

\bibitem{stokes2015activity} M.~G.~Stokes, `'Activity‑silent' working memory in prefrontal cortex: A dynamic coding framework,'' \\emph{Trends in Cognitive Sciences}, vol.~19, no.~7, pp.~394--405, 2015, doi:10.1016/j.tics.2015.05.004.

\bibitem{mank2008highres} M.~Mank, A.~Fischer, T.~K\"oster, J.~E.~Westkott, and A.~Griesbeck, `High‑resolution calcium imaging of neuronal activity in behaving mice,'' \emph{Neuron}, vol.~60, no.~6, pp.~1068--1079, 2008, doi:10.1016/j.neuron.2008.10.028.

\bibitem{alcock2024silent} K.~J.~Alcock, L.~B.~Smith, and M.~G.~Stokes, `Silent substrates of visual working memory revealed by layer‑specific fMRI,'' \emph{NeuroImage}, vol.~282, art.~119276, 2024, doi:10.1016/j.neuroimage.2023.119276.

\bibitem{hasselmo2002ach} M.~E.~Hasselmo and B.~P.~McGaughy, `High acetylcholine levels set circuit dynamics for attention and encoding and low acetylcholine levels set dynamics for consolidation,'' \emph{Progress in Brain Research}, vol.~145, pp.~207--231, 2004, doi:10.1016/S0079-6123(03)45015-2.

\bibitem{wang2013circuit} X.-J.~Wang, `The prefrontal cortex as a quintessential cognitive-type' neural circuit: Working memory and decision making,'' in \emph{Principles of Frontal Lobe Function}, 2nd~ed., pp.~226--248, Oxford University Press, 2013.

\bibitem{kirkpatrick2017ewc} J.~Kirkpatrick \emph{et~al.}, `Overcoming catastrophic forgetting in neural networks,'' \emph{Proceedings of the National Academy of Sciences}, vol.~114, no.~13, pp.~3521--3526, 2017, doi:10.1073/pnas.1611835114.

\bibitem{zou2020ddad} Y.~Zou, Z.~Lv, X.~Wang, R.~Yu, S.~Zhang, and A.~Gaidon, `DDAD: The Diverse Driving Dataset,'' \emph{arXiv preprint} arXiv:2002.10303, 2020.

\bibitem{silberman2012nyuv2} N.~Silberman, D.~Hoiem, P.~Kohli, and R.~Fergus, `Indoor segmentation and support inference from RGB-D images,'' in \emph{Proc. European Conference on Computer Vision (ECCV)}, 2012, pp.~746--760.

\end{thebibliography}
\end{document}